\documentclass{article}
\usepackage[preprint]{neurips_2025}

\usepackage{embedall}
\usepackage{microtype}
\usepackage{times}
\usepackage{helvet}
\usepackage{courier}
\usepackage{algpseudocode}
\usepackage[hyphens]{url}
\usepackage{graphicx}
\usepackage{algorithm}
\usepackage{adjustbox}
\usepackage{amsmath}
\usepackage{float}
\usepackage{booktabs}
\usepackage{soul, color}
\usepackage{multirow}
\usepackage{tabularx}
\usepackage{makecell}
\usepackage{ragged2e}
\usepackage{xcolor}
\usepackage[table]{xcolor}
\definecolor{verylightgray}{gray}{0.95}
\definecolor{lightpink}{RGB}{255,182,193}
\newcolumntype{g}{>{\columncolor{verylightgray}}c}
\newcolumntype{p}{>{\columncolor{lightpink}}c}
\usepackage{newfloat}
\usepackage{listings}
\floatstyle{ruled}
\newfloat{listing}{tb}{lst}{}
\floatname{listing}{Listing}

\makeatletter
\renewcommand{\@noticestring}{Accepted to the NeurIPS 2025 Workshop: UrbanAI — Harnessing Artificial Intelligence for Smart Cities.}
\makeatother

\title{AI-Driven Forecasting and Monitoring of Urban Water System}

\author{
  Qiming Guo\textsuperscript{1,2}, Bishal Khatri\textsuperscript{1}, Hua Zhang\textsuperscript{1}, Wenlu Wang\textsuperscript{1,2} \\
  \textsuperscript{1}Texas A\&M University - Corpus Christi, USA,
  \textsuperscript{2}The AIII Lab, USA, \\
  qguo2@islander.tamucc.edu, bkhatri1@islander.tamucc.edu\\
  hua.zhang@tamucc.edu, wenlu.wang@tamucc.edu
}

\begin{document}

\maketitle
\begin{abstract}
Underground water and wastewater pipelines are vital for city operations but plagued by anomalies like leaks and infiltrations, causing substantial water loss, environmental damage, and high repair costs. Conventional manual inspections lack efficiency, while dense sensor deployments are prohibitively expensive. In recent years, artificial intelligence has advanced rapidly and is increasingly applied to urban infrastructure. In this research, we propose an integrated AI and remote-sensor framework to address the challenge of leak detection in underground water pipelines, through deploying a sparse set of remote sensors to capture real-time flow and depth data, paired with HydroNet—a dedicated model utilizing pipeline attributes (e.g., material, diameter, slope) in a directed graph for higher-precision modeling. Evaluations on a real-world campus wastewater network dataset demonstrate that our system collects effective spatio-temporal hydraulic data, enabling HydroNet to outperform advanced baselines. This integration of edge-aware message passing with hydraulic simulations enables accurate network-wide predictions from limited sensor deployments. We envision that this approach can be effectively extended to a wide range of underground water pipeline networks.
\end{abstract}

\section{Introduction}

Urban underground water and wastewater pipelines are critical infrastructure for city operations, enabling the efficient transport of water and wastewater. However, these systems are costly to construct and maintain, and as they age, they become increasingly vulnerable to fractures and faulty connections. Such anomalies—including leaks, infiltrations, and blockages—result in substantial water loss, environmental contamination, and rising repair costs. Given these challenges, accurate detection of faulty connections and fractures is essential. Traditional approaches, however, are limited: manual inspections often suffer from delays and incomplete coverage, while dense sensor deployments remain economically prohibitive due to high installation and maintenance costs. Other assessment techniques, such as closed-circuit television (CCTV) surveys and smoke testing \cite{beheshti2019detection}, offer more comprehensive diagnostics but are both costly and labor-intensive

In recent years, AI has advanced rapidly, and developing dedicated AI techniques for urban applications offers a promising solution. To address this challenge, we present an AI-enhanced monitoring system that deploys sparse sensors at manholes to collect real-world water (RWW) data such as inflow and depth, augmented by hydraulic simulations for full network coverage. This RWW data then feeds into HydroNet, our spatiotemporal graph neural network that represents the water and wastewater network as a directed graph—nodes as manholes, edges as pipes with physical attributes (diameter, slope, material). By incorporating these pipeline properties directly into message passing, HydroNet learns normal hydraulic patterns and flags anomalies through deviation detection. We deployed this integrated AI and remote-sensor framework on a campus network and collected over a year of real-world data for modeling and evaluation. Experimental results demonstrate that the system effectively captures spatiotemporal hydraulic data and enables HydroNet to achieve high precision, with the best MAE of 0.0085 ft for depth and 0.0038 cfs for flow. These results confirm that the system successfully provides data with strong predictive utility, supporting asynchronous forecasting and anomaly detection in pipeline flow. This work represents a successful practice of AI for urban underground water monitoring.

\begin{figure*}[!h]
  \centering
  \includegraphics[width=1\linewidth]{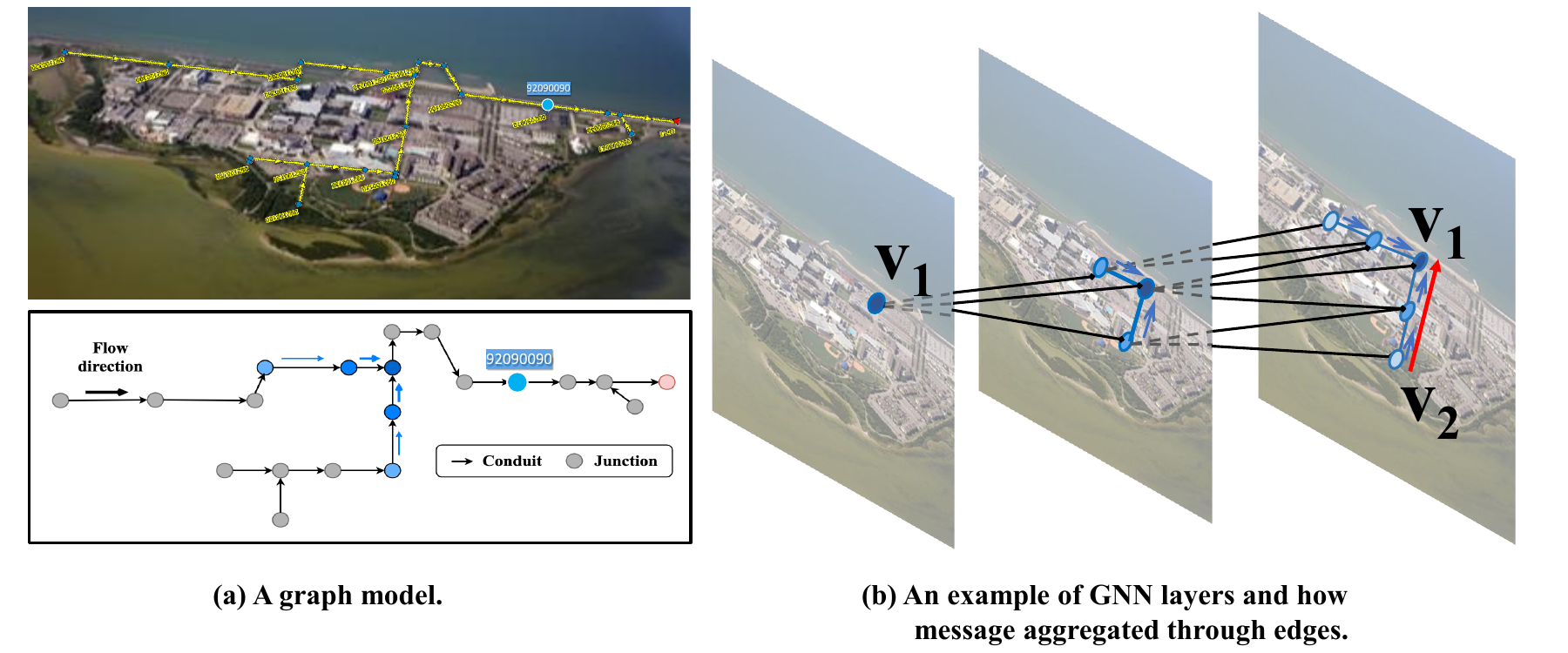}
\caption{Graph representation of the campus sewer network. Nodes represent manholes with flow and depth measurements, while directed edges represent pipes with physical attributes.}
  \label{fig:water-gnn}
\end{figure*}

\section{Methodology}

Our system operates in three interconnected stages: first, sparse remote sensors collect real-time flow and depth data from key network locations, augmented by hydraulic simulations for full coverage; second, this data is fused with pipeline attributes in a graph-structured format for spatio-temporal modeling via HydroNet; and third, the model learns to accurately predict normal hydraulic patterns. This end-to-end pipeline balances cost-efficiency with accuracy, leveraging physical domain knowledge to enhance AI predictions. We detail each component below.

\begin{figure*}[b!]
    \centering
    \includegraphics[width=1\textwidth]{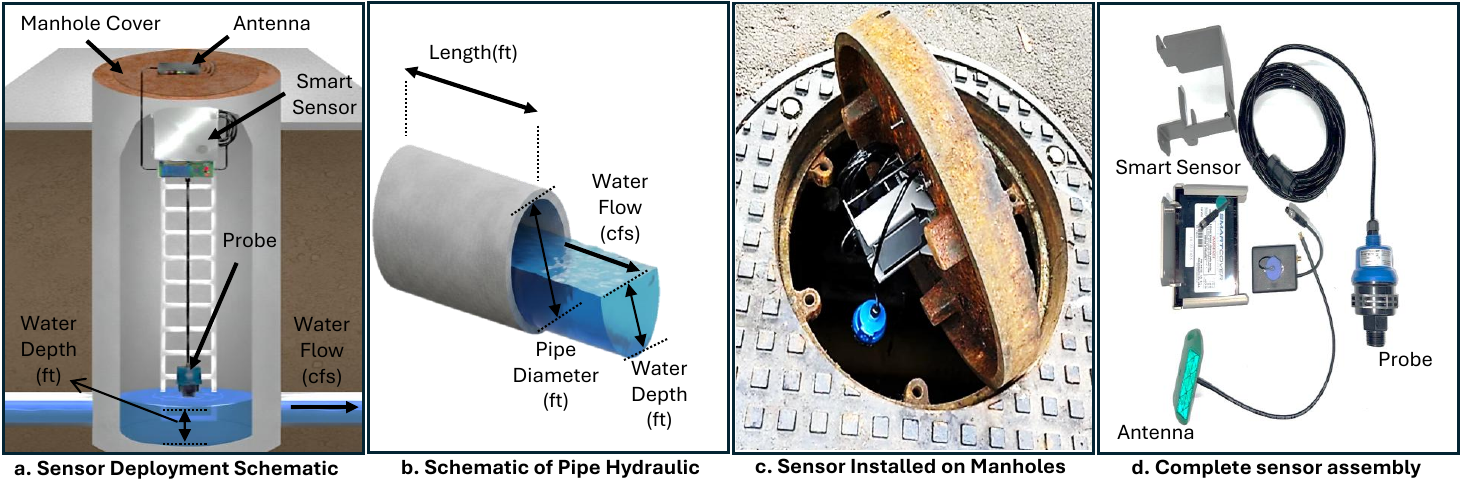}
    \caption{Schematic of the remote sensor system and its deployment configuration.}
    \label{fig:sensor}
\end{figure*}

\subsection{Remote Sensing System}

We deployed SmartCover sensors Figure~\ref{fig:sensor}(a) in the underground wastewater system of a university campus. 
The network is modeled as a directed graph $G=(V, E)$, where $V$ denotes manholes (nodes) and $E$ represents pipes (edges) for downstream message passing (Figure~\ref{fig:water-gnn}). This structure enables propagation of anomaly impacts, e.g., from an upstream leak to downstream nodes.

Sensors are affixed to manhole interiors, as shown in Figure~\ref{fig:sensor}(c), avoiding pipeline entry and minimizing installation risks. To optimize costs, each sensor measures only flow velocity and water depth, as shown in Figure~\ref{fig:sensor}(b). Data are transmitted in real time via antenna, as shown in Figure~\ref{fig:sensor}(d), eliminating manual retrieval. The devices are battery powered with a two-year lifespan, supporting sustained monitoring without frequent maintenance.

\subsection{RWW Dataset: Collection and Characterization}

% SmartCover sensors (Badger Meter, CA, USA) were installed at selected manholes, with PCSWMM hydraulic model simulating the remaining nodes to ensure full coverage across all 23 nodes.

\begin{figure}[b!]
\centering
\includegraphics[width=1\linewidth]{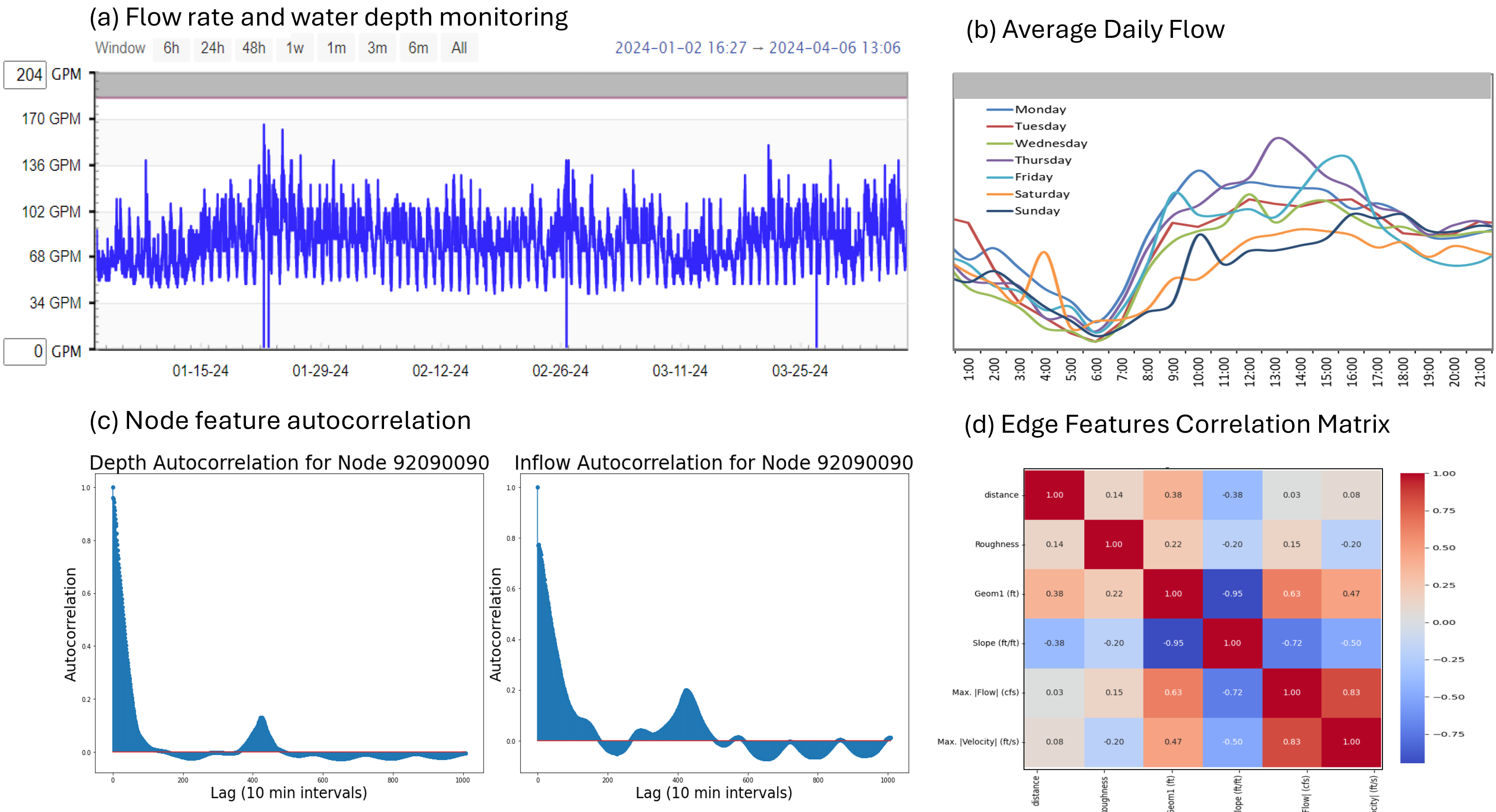}
\caption{Analysis of the RWW dataset features: (a) Time-series visualization of aggregated flow rate across the network, highlighting periodic variations; (b) Average daily flow patterns by day of the week; (c) Example autocorrelation functions for depth and inflow at node 92090090, highlighted in Figure~\ref{fig:water-gnn}(a); (d) Correlation matrix of edge features}
\label{fig:analyze}
\end{figure}

We collect Real World Water (RWW) data from a wastewater network with 22 vitrified clay pipes and 23 nodes (22 manholes plus 1 outlet, with invert elevations -1.69 to 0.56 m; Figure~\ref{fig:water-gnn}). SmartCover sensors were deployed at 5 selected manholes, recording water depth and flow rate at 10-minute intervals from October 1, 2023, to January 31, 2024, as shown in Figure~\ref{fig:analyze}(a). A calibrated PCSWMM hydraulic model provided data for the remaining 18 nodes at the same temporal resolution, demonstrating effective sparse sensing augmented by simulation. Node features include time-series data (17,706 steps) of water depth and flow rate; edge features include static attributes as shown in Table~\ref{tab:features}.

%such as length, roughness, diameter (Geom1), slope, GIS length, max flow, max velocity, max / full flow, and max / full depth. 

\begin{table}[h!]
\caption{{Node (time-series) and edge (static) features in the RWW dataset.}}
\label{tab:features}
\centering
\small
\renewcommand{\arraystretch}{1.2}
\begin{tabular}{>{\bfseries}m{1cm} m{8cm}}
\toprule
Node & Water depth, Flow rate \\
\midrule
Edge & Length, Roughness, Diameter (Geom1), Slope, GIS Length, Max Flow, Max Velocity, Max / Full Flow, Max / Full Depth \\
\bottomrule
\end{tabular}
\end{table}

This data monitors hydraulic behavior, capturing daily and seasonal variations in hydraulic flow, as illustrated in the sewer pipe hydraulics schematic in Figure~\ref{fig:sensor}(b)
. Site-collected features include pipe length, diameter, and slopes. The dataset shows consistent oscillatory patterns from short- and long-term flow variations, influenced by operational schedules, diurnal cycles, or external factors (e.g., rainfall). Leveraging this data refines predictive models for anomaly detection, peak load forecasting, and water infrastructure management optimization.

We used the autocorrelation function (ACF) to analyze water depth and flow rate across nodes, observing stability and temporal evolution, as shown in Figure~\ref{fig:analyze}(c) for node 92090090 in Figure~\ref{fig:water-gnn}(a). Most nodes show high short-term autocorrelation (first 100-200 hours), indicating initial oscillations from environmental or random factors, followed by cyclical variations after 2-3 days, signaling periodicity. Flow rate data exhibits similar trends with slightly more early variability than water depth. The edge features correlation heatmap in Figure~\ref{fig:analyze}(d) reveals significant relationships, including strong negative correlations (-0.95 between slope and diameter (Geom1); -0.72 between max flow and slope) and positive correlation (0.83 between max flow and velocity), highlighting diverse hydraulic behaviors. Figure~\ref{fig:analyze}(b) shows average daily flow patterns by day of the week, with daytime peaks from increased activity, early-morning troughs, and fluctuations reflecting varying consumption habits influenced by operational or community activities. This RWW dataset has also been used for modeling and benchmarking in \citep{guo2025efficient,guo2024hydronet}, and is publicly available at \url{https://github.com/VV123/STEPS}.

\subsection{HydroNet Modeling}
HydroNet processes graph-structured time-series via two ST-MPNN blocks and an output layer. Each block combines gated temporal convolutions for capturing time dependencies with a custom MPNN for spatial propagation. The temporal component processes a lookback window of $L$ time steps (set to 12 in our experiments) to learn patterns in the time series. For spatial modeling, we incorporate the edge features from Table~\ref{tab:features} as vectors $\mathbf{a}_{ij}$ for each edge $(i,j)$. These attributes are transformed through a learnable embedding matrix $\mathbf{W}_a$ and integrated into the message passing:
\[
h_i^{(t)} = \text{TempConv}(x_i^{(t-L:t)}), \quad
m_{ij} = f_\text{message}(h_i^{(t)}, \mathbf{W}_a \mathbf{a}_{ij})
\]
\[
h_j' = f_\text{update}\left(h_j^{(t)}, \sum_{i \in \mathcal{N}(j)} m_{ij}\right)
\]
where $f_\text{message}$ and $f_\text{update}$ are learnable functions. The nine edge features in $\mathbf{a}_{ij}$ are jointly embedded via $\mathbf{W}_a$ to learn their relative importance. This design enables HydroNet to learn complex spatio-temporal patterns while respecting the physical constraints encoded in pipe attributes.

\section{Evaluation}
We evaluate the predictive modeling capability of our AI-powered sensor network system using the Real World Water (RWW) dataset. Both advanced baseline models and our proposed HydroNet were trained and tested for comparison. The dataset was split into training, validation, and testing sets in a 7:1:2 ratio, with a lookback window of 12 time steps to forecast the subsequent 12 steps. All experiments were conducted on a high-performance server equipped with NVIDIA A6000 GPUs (48 GB VRAM). Early stopping based on validation loss was applied to ensure fair comparison. Table~\ref{tab:rww_results} reports the forecasting performance for water depth and flow rate. HydroNet consistently outperforms baselines across all metrics, demonstrating the effectiveness of edge-attribute-aware message passing in capturing spatiotemporal hydraulic dynamics.

\begin{table*}[h!]
\caption{Performance on the RWW dataset.}
\label{tab:rww_results}
\centering
\small
\setlength{\tabcolsep}{3.5pt}
\renewcommand{\arraystretch}{1.1}
\begin{tabular}{l@{\hspace{4pt}}ccc@{\hspace{10pt}}ccc}
\toprule
& \multicolumn{3}{c}{\textbf{Depth(ft)}} & \multicolumn{3}{c}{\textbf{Flow(cfs)}} \\
\cmidrule(lr){2-4} \cmidrule(lr){5-7}
\textbf{Method} & MAE & RMSE & MAPE & MAE & RMSE & MAPE \\
\midrule
CaST~\citep{xia2024deciphering}       & 0.0186 & 0.0298 & 0.0747 & 0.0077 & 0.0138 & 0.0358 \\
GMAN~\citep{zheng2020gman}            & 0.0186 & 0.0140 & 0.0130 & 0.0168 & 0.0181 & 0.1255 \\
ST-SSL~\citep{ji2023spatio}           & 0.0196 & 0.0230 & 0.0273 & 0.0150 & 0.0322 & 0.1313 \\
STG-MAMBA~\citep{li2024stg}           & 0.0176 & 0.0296 & 0.0120 & 0.0098 & 0.0166 & 0.1373 \\
STGCN~\citep{stgcn}                   & 0.0123 & 0.0324 & 0.0657 & 0.0066 & 0.0158 & 0.0709 \\
\midrule
\textbf{HydroNet}                     & 0.0085 & 0.0178 & 0.0454 & 0.0038 & 0.0094 & 0.0408 \\
\bottomrule
\end{tabular}
\end{table*}

\section{Conclusion}
We introduced an AI-powered monitoring framework for urban underground wastewater pipelines that combines sparse sensing with hydraulic simulations for cost-effective data collection. The collected measurements are modeled by HydroNet, an edge-aware spatiotemporal graph neural network that incorporates pipeline attributes to improve predictive accuracy. Experiments on a real-world campus network show that the framework achieves highly precise flow and depth forecasts, providing strong predictive utility. These forecasts provide a robust foundation for downstream tasks such as anomaly detection, offering a scalable, data-driven solution to enhance urban water infrastructure security. 

\section*{Limitations}

While our approach shows promise, it has several limitations. The system was tested on a campus-scale network, which may not fully represent larger urban environments with more complex topologies or variable conditions. Data collection relied on specific sensors, and performance could vary with different hardware or environmental factors like extreme weather. The models assume accurate pipeline attribute data, which may not always be available or up-to-date in real-world scenarios. Additionally, the high-level focus means that fine-grained optimizations for specific anomaly types are not explored, and further validation on diverse datasets is needed to ensure generalizability.

\section*{Ethics Statement}

This research adheres to ethical standards in AI and environmental engineering. Data collection was conducted on a university campus with proper permissions, ensuring no disruption to operations or privacy concerns, as no personal data was involved. The system's goal of anomaly detection promotes water conservation and environmental protection, aligning with sustainable development objectives. However, we acknowledge potential dual-use risks, such as misuse of sensor data for unauthorized surveillance, and emphasize the need for secure data handling. Bias in modeling could arise from incomplete datasets, so we recommend ongoing audits. Overall, this work aims to benefit society by enhancing infrastructure resilience without compromising ethical principles.

\section*{Acknowledgements}

This work was supported by the National Science Foundation (NSF) under Award No. 2318641.

\bibliographystyle{plainnat}
\bibliography{references}

\begin{thebibliography}{8}
\providecommand{\natexlab}[1]{#1}
\providecommand{\url}[1]{\texttt{#1}}
\expandafter\ifx\csname urlstyle\endcsname\relax
  \providecommand{\doi}[1]{doi: #1}\else
  \providecommand{\doi}{doi: \begingroup \urlstyle{rm}\Url}\fi

\bibitem[Beheshti and S{\ae}grov(2019)]{beheshti2019detection}
Maryam Beheshti and Sveinung S{\ae}grov.
\newblock Detection of extraneous water ingress into the sewer system using tandem methods--a case study in trondheim city.
\newblock \emph{Water Science and Technology}, 79\penalty0 (2):\penalty0 231--239, 2019.

\bibitem[Guo and Wang(2024)]{guo2024hydronet}
Qiming Guo and Wenlu Wang.
\newblock Hydronet: A spatio-temporal graph neural network for modeling hydraulic dependencies in urban wastewater systems.
\newblock In \emph{Proceedings of the 32nd ACM International Conference on Advances in Geographic Information Systems}, pages 717--718, 2024.

\bibitem[Guo et~al.(2025)Guo, Pan, Zhang, and Wang]{guo2025efficient}
Qiming Guo, Chen Pan, Hua Zhang, and Wenlu Wang.
\newblock Efficient unlearning for spatio-temporal graph (student abstract).
\newblock In \emph{Proceedings of the AAAI Conference on Artificial Intelligence}, volume~39, pages 29382--29384, 2025.

\bibitem[Ji et~al.(2023)Ji, Wang, Huang, Wu, Xu, Wu, Zhang, and Zheng]{ji2023spatio}
Jiahao Ji, Jingyuan Wang, Chao Huang, Junjie Wu, Boren Xu, Zhenhe Wu, Junbo Zhang, and Yu~Zheng.
\newblock Spatio-temporal self-supervised learning for traffic flow prediction.
\newblock In \emph{Proceedings of the AAAI Conference on Artificial Intelligence}, volume~37, pages 4356--4364, 2023.

\bibitem[Li et~al.(2024)Li, Wang, Zhang, and Coster]{li2024stg}
Lincan Li, Hanchen Wang, Wenjie Zhang, and Adelle Coster.
\newblock Stg-mamba: Spatial-temporal graph learning via selective state space model.
\newblock \emph{arXiv preprint arXiv:2403.12418}, 2024.

\bibitem[Xia et~al.(2024)Xia, Liang, Wen, Liu, Wang, Zhou, and Zimmermann]{xia2024deciphering}
Yutong Xia, Yuxuan Liang, Haomin Wen, Xu~Liu, Kun Wang, Zhengyang Zhou, and Roger Zimmermann.
\newblock Deciphering spatio-temporal graph forecasting: A causal lens and treatment.
\newblock \emph{Advances in Neural Information Processing Systems}, 36, 2024.

\bibitem[Yu et~al.(2017)Yu, Yin, and Zhu]{stgcn}
Bing Yu, Haoteng Yin, and Zhanxing Zhu.
\newblock Spatio-temporal graph convolutional networks: A deep learning framework for traffic forecasting.
\newblock \emph{arXiv preprint arXiv:1709.04875}, 2017.
\newblock URL \url{https://arxiv.org/pdf/1709.04875}.

\bibitem[Zheng et~al.(2020)Zheng, Fan, Wang, and Qi]{zheng2020gman}
Chuanpan Zheng, Xiaoliang Fan, Cheng Wang, and Jun Qi.
\newblock Gman: A graph multi-attention network for traffic prediction.
\newblock In \emph{Proceedings of the AAAI Conference on Artificial Intelligence}, volume~34, pages 1234--1241, 2020.

\end{thebibliography}

\end{document}